\documentclass{article}
\usepackage{booktabs}
\usepackage{amsmath}
\usepackage{mdframed}
\usepackage{geometry}
\usepackage{pifont}
\usepackage{xcolor}
\usepackage{booktabs,tabularx}
\usepackage{multirow} 
\usepackage{enumitem}
\usepackage{authblk}

\usepackage{amsmath}
\usepackage{amssymb}
\usepackage{graphicx}
\usepackage{cite}
\usepackage{color}

\usepackage{microtype}
\usepackage{xcolor}
\usepackage{subfig}
\usepackage[ruled,vlined,linesnumbered]{algorithm2e}
\SetNlSty{bfseries}{\color{black}}{}
\SetKwInput{KwInputData}{Input data}
\SetKwInput{KwActors}{Actors}
\SetKwInput{KwRoutines}{Routines}
\SetKwInput{KwHyperparameter}{Hyperparameter}
\SetKwFor{ForParallel}{for}{do in parallel}{end}

\usepackage{hyperref}
\hypersetup{
    colorlinks=true,
    linkcolor=blue,
    filecolor=magenta,      
    urlcolor=cyan,
    pdftitle={Overleaf Example},
    pdfpagemode=FullScreen,
    }

\newcommand{\proj}{\textit{AFˆ2 Framework}\xspace}

\newcommand{\cprojspec}[1]{{\color{brown} #1}}
\newcommand{\cdataprov}[1]{{\color{purple} #1}}
\newcommand{\cprepro}[1]{{\color{purple} #1}}
\newcommand{\ctrain}[1]{{\color{teal} #1}}
\newcommand{\cpostpro}[1]{{\color{blue} #1}}

\title{Towards an Accountable and Reproducible Federated Learning:\\ A FactSheets Approach}
\author[1]{Nathalie Baracaldo}
\author[1]{Ali Anwar}
\author[1]{Mark Purcell} 
\author[1]{Ambrish Rawat}
\author[1]{Mathieu Sinn\footnote{work done while at IBM}}
\author[1]{\\Bashar Altakrouri$^*$}
\author[2]{Dian Balta}
\author[2]{Mahdi Sellami} 
\author[2]{Peter Kuhn}
\author[2]{\\Ulrich Schopp}
\author[2]{ Matthias Buchinger}
\affil[1]{IBM Research}
\affil[2]{fortiss}

\date{}

\begin{document}
\maketitle
\begin{abstract}

Federated Learning (FL) is a novel paradigm for the shared training of models based on decentralized and private data. With respect to ethical guidelines, FL is promising regarding privacy, but needs to excel vis-à-vis transparency and trustworthiness. In particular, FL has to address the accountability of the parties involved and their adherence to rules, law and principles. We introduce \proj, where we instrument FL with accountability by fusing verifiable claims with tamper-evident facts, into reproducible arguments. We build on AI FactSheets for instilling transparency and trustworthiness into the AI lifecycle and expand it in order to incorporate dynamic and nested facts, as well as complex model compositions in FL. Based on our approach, an auditor can validate, reproduce and certify a FL process. This can be directly applied in practice to address the challenges of AI engineering and ethics.
\end{abstract}

\section{Introduction}

Artificial Intelligence (AI) and machine learning (ML) have the potential to enhance a wide range of business, industrial and social applications. Nonetheless, these technologies come with grand ethical challenges. Recent research on ethical guidelines around the world \cite{jobin_global_2019} shows that the core ethical concepts and principles for AI and ML are still embryonic and are yet to be operationalized in practice \cite{leslie_david_2019_3240529}. Other research suggests that existing concepts and principles might be insufficient \cite{whittlestone_role_2019, mittelstadt_principles_2019}. 

Two ethical concepts are of particular relevance from an engineering perspective, namely \textit{privacy} and \textit{accountability}. On one hand, privacy addresses how to handle data that is personal or can be attributed to a person. To this end, privacy legislation, such as the GDPR or HIPAA, places strict limits on the amount of personal data that can be collected and transmitted to a central location. Such limits conflict with the needs of ML-driven applications, which mostly rely on the availability of data. On the other hand, accountability represents a relationship between actors involved in ML, by providing the means as well as mechanisms for transparency and control. Accountability is particularly critical for the acceptance of ML in highly regulated environments, where auditing and fairness is paramount (cf. e.g. \cite{wilson2021building} ).

\textit{Federated Learning} (FL)\cite{fedavgMahan}, is a privacy-by-design engineering approach created to overcome the privacy challenges present in the traditional ML training process.
 In contrast to the existing centralized ML approaches that require the collection of training data at a single place, in FL the training data \textit{always} remains with their rightful owner.
 FL is designed to allow multiple \textit{parties}, each possessing their respective training data, to collaboratively build a shared model, without exchanging or transmitting this training data.
 For example, multiple governmental agencies (parties) may collect information about citizens that is private and cannot be transmitted to a central place due to regulations, yet the agencies and public may benefit from the creation of an ML model capable of speeding up processes.
 FL enables these agencies to collaboratively train the model, without explicitly sharing their data.
 
 While only recently introduced in 2019, FL has been embraced by a wide range of industries and has appeared as one of the five technologies for insights-driven businesses in the 2021 Forrester report{\footnote{\url{https://www.forrester.com/go?objectid=RES163520}}} with companies such as IBM and Nvidia developing products in the space. Common use cases of FL include building a shared model between numerous smart devices for next work prediction~\cite{apple,google}, spell check during typing, multiple sensors for event recognition, multiple financial organizations recognizing money laundering or predicting credit risk~\cite{webank}, or various healthcare applications for diagnosis~\cite{rieke2020future}.


Although promising in terms of addressing privacy, FL has yet to address \textit{accountability}. For instance, the evidence for FL adherence to governance, risk and compliance obligations is crucial to monitoring and auditing the fulfilment of obligations among multiple parties.
The distributed aspect of FL, where multiple multiple parties are involved along the FL process, presents challenges for accountability, especially if the parties involved are separate companies or government entities.
For instance, the users of a FL-powered credit score application expect it to perform according to its published specification. Specified rules should involve protection of privacy as well as results free from gender or racial bias. But what evidence do users have to verify that the application does indeed perform as expected? What are the mechanisms for review, audit and enforcement or rules that are available to users?

In this paper, we design FL as an \textit{accountable system} \cite{kacianka2021designing}, drawing on some accountability requirements specified for traditional ML: prove adherence to governance, risk and compliance obligations and allow for monitoring and auditing the fulfilment of the latter among multiple parties. In traditional ML, \textit{ FactSheets}~\cite{arnold2019factsheets} have been proposed as a way to enable such accountability by recording ``facts" related to the overall ML pipeline, in terms of performance and robustness guarantees, safety and security as well as a lineage of data and model versions. In contrast, engineering accountable systems in FL settings is even more challenging given the distributed nature of the system and the potential lack of trust among different parties involved in the training life-cycle. In fact, the distributed and privacy-by-design nature of FL make it non-transparent. Consequently, accountability for the FL process requires additional provisions that enable different entities in the system to verify their behavior, which is not the case in traditional ML. 

Our proposed framework, the \textit{Accountable FL FactSheet Framework} (\proj), addresses the engineering challenges for enabling accountability in FL. Our framework provides a formalized definition of accountability represented as a set of verifiable, undeniable, certifiable and tamper-evident claims. Moreover, \proj adapts the use of FactSheets as a practical approach to operationalizing accountability in practice and providing a single view of the information collected. We also provide and discuss a practical implementation for \proj. Finally, we apply our approach to an exemplary scenario from the government domain, namely for online citizen participation. Here, FL is used to build a shared text-based ML classification model, between multiple participating cities, to classify participation ideas into multiple categories.




\section{Background and Related Work}
\label{sec:background}

\subsection{Federated Learning for Privacy-Preserving ML}
\begin{figure}[h]
    \centering
    \includegraphics[width=0.6\textwidth]{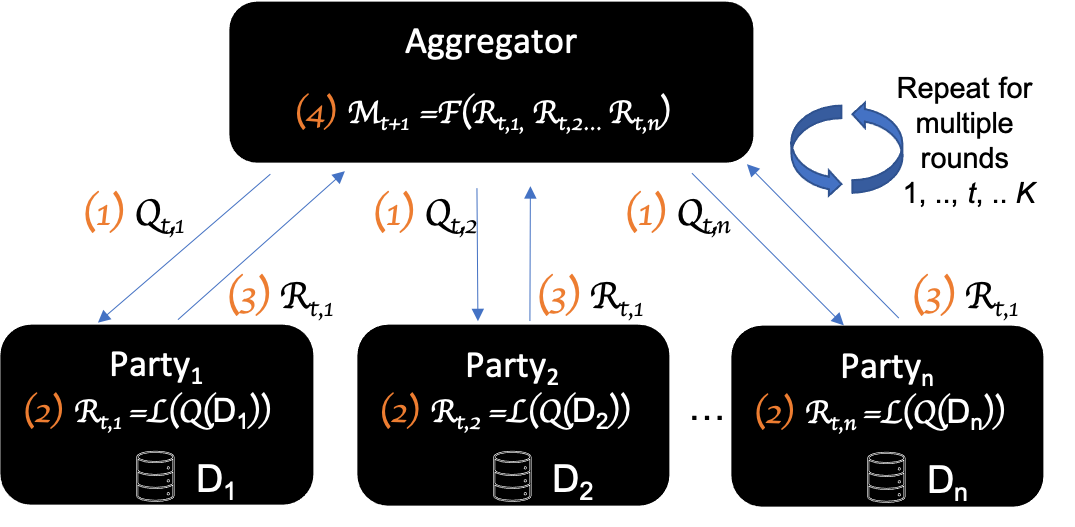}
    \caption{Federated learning process}
    \label{fig:basics}
\end{figure}

FL is a paradigm for training ML models on distributed data without having to gather the training data at a central place~\cite{fedavgMahan,kairouz2021advances}. FL is applicable to a broad family of ML models, including Neural Networks (NN), Gradient-Boosted Decision Trees, and Support Vector Machines.
Figure \ref{fig:basics} presents an overview of the FL process: The \textit{aggregator} $\mathcal{A}$ orchestrates the collaborative training of an \textit{ML model}
among a set of \textit{parties} $P_1, P_2, \dots, P_n$. 
Each of these parties holds their own private training \textit{datasets}, denoted $D_1, D_2, \dots, D_n$ that cannot be transmitted during the learning process.

Initially, a common \textit{model specification} and FL algorithm are agreed upon. A plethora of such algorithms are available for different types of ML models, e.g.~\cite{fedavg,pfmn,xgboostFLibm,zeno}.
These FL algorithms are distributed and can be separated into the \textit{fusion algorithm} $\mathcal{F}$ run by the aggregator, and \textit{local training algorithm} $\mathcal{L}$, run by the parties. Additionally, specific methods for local data pre-processing may be prescribed. With this information, the federation is set up and parties register with the aggregator.


Next, the aggregator starts orchestrating the training process according to the specified \textit{fusion algorithm} $\mathcal{F}$, which defines the queries issued to the parties and how the parties' replies are combined to advance the training process. Typically, the federated training process requires sending multiple queries to the parties before arriving to a suitable model, known as the number of \textit{rounds} $t=1, 2, \dots, K$.

The query in round $t$ to a party $P_i$, is denoted by $Q_{t,i}$.
While some fusion algorithms query all parties \cite{fedavg,zeno}, others only query a subset of the parties in each round \cite{tifl}, often depending on the total number of parties involved in the federation.
In step (1) of Figure \ref{fig:basics}, in round $t$, the aggregator sends a query $Q_{t,i}$ to each party $i$.
The structure of queries depends on the type of ML model, e.g., for NNs
a query contains typically an initial NN model and requests parties to continue training it on their local data, before sending the trained model's weights back to the aggregator.
If a decision tree is trained, the query may request parties to send to the aggregator the counts for a particular feature. 

Upon receiving a query $Q_{t,i}$, party $P_i$ uses its dataset $D_i$ to answer the query by running a local training function $\mathcal{L}$ (step (2) in Figure \ref{fig:basics}).
As a result of executing $\mathcal{L}$, a reply $R_{t,i}$ is sent to the aggregator (step (3)). 
In the NN case, the NN included in the query is used to begin local training on $D_i$, parameterized by a set of hyperparameters $\eta_{pi}$ that may include the learning rate and the number of local training epochs.

Meanwhile the aggregator waits for a pre-specified time for parties to reply.
After this time has elapsed, the aggregator may verify if there is a \textit{quorum} (i.e.~required minimum number of replies) to proceed with the training process and, if this is the case, proceed with aggregating all received replies and creating a new model as shown in step (4) of Figure \ref{fig:basics}. We denote this new model as $\mathcal{M}_{t+1}$.
This process repeats until the pre-specified number of rounds $K$ has been completed.

The above process is generic and there are numerous algorithms that can embody $\mathcal{F}$ and $\mathcal{L}$.
Variations include how replies are aggregated, different weighting of replies in the aggregation (with the extreme case of certain replies being completely removed from the aggregation). Some algorithms adapt the hyperparameters of $\mathcal{F}$ or $\mathcal{L}$ throughout the training rounds.
Other algorithms may add differential privacy \cite{hybridone} or encrypt replies, e.g.~via multi-party computation techniques \cite{hybridalpha}.
Some flavors of $\mathcal{F}$ make use of hold-out data, e.g.~to determine potentially malicious replies and exclude them from the aggregation process \cite{zeno}, or to evaluate early stopping criteria.
The hold-set can also be used to evaluate early termination (e.g., if the accuracy of the model is greater than 95\% for the hold-out set, stop the training).

Most work on federated learning assumes that the data at each of the parties is readily available for training.
This assumption may hold for some application, such as mobile phone ones. However, in the enterprise setting, this is typically not the case.
For this reason, prior to the training process, a pre-processing phase needs to occur.
Here, all the datasets need to be transformed to ensure data is in the same format: it has been properly normalized and column order is the same in all entities.
As part of pre-processing, each party may also run algorithms to reduce disparate impact and provide fair models \cite{abay}. 
The pre-processing process is extremely relevant to ensure that the final model has good performance as well as the reproducibility of the overall process and its results is.   

In summary, this plethora of algorithmic configurations and pre-processing steps allow for precise adjustment of FL to the requirements of a particular application, while at the same time the process of auditing \cite{raji2020closing} proves even more complex and bears additional challenges for the trustworthiness of the overall system (cf. e.g. \cite{wieringa2020account}) towards accountability.


\subsection{Accountability and Reproducibility for Machine Learning Systems}
Accountability has been in the research and practical focus of various disciplines and actors with emphasis on social, technological and legal aspects (cf. e.g. \cite{kacianka2021designing}). 


An operationalization of accountability involves studying the applied ML algorithms \cite{wieringa2020account} as well as particular technology components applied as well as data \cite{toreini2020relationship}, and should involve a structured process that results in keeping corresponding records of the evidence generated according to predefined requirements \cite{raji2020closing}. We suggest a way to operationalize accountability through AI FactSheets.

The purpose of AI FactSheets is to provide transparency and instill trust into AI services \cite{arnold2019factsheets}. The idea is that providers of AI services shall explicitly describe -- in a so-called FactSheet -- all relevant attributes of an AI service, in particular how it was created, trained and deployed. 

Similar to an external audit, an inspection of the FactSheets allows consumers to make an informed decision whether or not to use the service. ``Consumers'' could be either end users, or businesses that consider using the AI services in downstream applications\footnote{For more background on related ideas, such as model cards, data sheets, software auditing and certification, we refer to \cite{arnold2019factsheets}}. While \cite{arnold2019factsheets} proposes AI FactSheets as a completely voluntary undertaking by AI service providers, it indicates that such documentation could evolve to be an industry requirement -- for certain industries or for certain types of AI services likely sooner than for others -- and potentially lead to a system of third-party certifications.

In regard to reproducibility, \textbf{factsheets} address four trust properties\cite{arnold2019factsheets}:
\begin{itemize}
    \item Basic performance and reliability;
    \item Safety (performance under dataset shifts, algorithmic fairness, explainability etc.);
    \item Security (robustness against adversarial attacks, malfunction etc.);
    \item Lineage (provenance of datasets, metadata, models, algorithms, hyperparameters etc.)
\end{itemize}

Addressing the requirements for factsheets to be auditable, which essentially requires an auditor to be able to reproduce all the reported facts, is arguably accomplished based on distributed, immutable ledgers for documenting the lineage and evolution of AI services built by multiple parties \cite{arnold2019factsheets}.

Applying directly existing approaches to generate Factsheets in FL is not as trivial given that the differences between the processes
followed to train ML models in a centralized and in FL settings.
Among the differences, we count the distributed nature of the system, the fact that parties do not share data and may not fully trust each other, among others.
In what follows, we introduced \proj to close this gap.
\section{Objectives of the \proj}
\label{sec:acc4FL}
With respect to tackling ethics challenges of AI by relying on principles \cite{whittlestone_role_2019,mittelstadt_principles_2019}, we study accountability from an engineering perspective. We focus on creating verifiable claims to support arguments that aim at trustworthy AI \cite{brundage_toward_2020}, by explaining the design and operation of a system, which allows for transparency, causality and ultimately control (e.g. blaming a  responsible actor for their actions and introducing countermeasures) \cite{kacianka2021designing}. An accountability argument should explain checks and tests performed as well as outcomes to ensure particular system properties. In particular, such an argument should be organized around particular claims (or goals) and supporting evidence about the system, i.e. it can be visualised as a tree, broken down into claims and subclaims (the interior is the reasoning) with evidence at the leaves \cite{bloomfield2020assurance}.  

To make the claims about ML verifiable, they should be formalized and their presence (as well as the supporting evidence) should allow for an audit trace \cite{raji2020closing}, which could be performed by the actor generating the claim \footnote{This is also referred to as \textit{internal audit} \cite{raji2020closing}} or a third party \footnote{This is also known as \textit{external audit} \cite{raji2020closing}}, not directly involved in the design and operation of a ML system. 

In the context of FL, verifying claims is more challenging, due to the distribution of the ML process and the decentralization of the generation of evidence. In this case, accountability should guaranteed  based on a third party than can judge upon the blame of misbehaviour between two parties  \cite{kunnemann2019automated, kunnemann2021accountability}, i.e. evidence provided to support the claims could be conflicting and conflicts should be resolvable by the third party. Consequently, the dimensions of accountable FL systems include \cite{yumerefendi_trust_2004,kusters_accountability:_2010, kacianka2021designing, kunnemann2021accountability}:

\begin{itemize}

\item \textit{Verifiable}: each step in the FL process must be supported with corresponding claims, transparent according to the aim of each claim, and evidence, that the step was conducted as specified and that the specification was sound. For instance, verification involves the need to reproduce a particular training step or check the correctness of a configuration file or of a communication protocol during the learning process. With regard to the properties of a FactSheet, a verifiable claim will involve the information about basic performance and reliability, safety and security of the FL.

\item \textit{Undeniable consent}: a FL process must be consistent with the knowledge of all actors\footnote{Here, please note that the knowledge includes all accessible information about the FL process configuration and generated model versions and potential metadata, but not the data itself}, each party must explicitly consent to the predefined specifications and the FL execution must be provable and non-repudiable. This would allow to assign directly responsibilities and to define the sufficient level of transparency, given that it is a prerequisite for consent.  

\item \textit{Auditable}: any misbehavior (based on attacks or faults) must by provable, based on corresponding claims supported by evidence, by an arbitrary third party. This would involve \textbf{reproducibility} of the FL process but also a rather comprehensive representation of the claims and supporting evidence in order to enable practical applicability towards certification.

\item \textit{Tamper-evident}: every interaction between actors must be recorded based on a predefined specification and any attempt to corrupt the recorded knowledge of all actors involved in the FL process incurs a high probability of detection. This dimension links to the lineage property of a factsheet.
\end{itemize}

In accordance with the presented dimensions, accountability of a FL system can be described as a set of \textit{verifiable}, \textit{undeniable}, \textit{auditable} and \textit{tamper-evident} claims that guarantee for 
(i) indicating faults in the distributed FL process that may have occurred when one of the entities in the system deviates from the protocols or there are byzantine errors,
(ii) reproducing the FL process,
and
(iii) allowing for better understanding of the ML models generated, their biases and when they can be applied. We note that in practice, each FL system will have to define accountability along the dimension depending on the context of its application. For instance, in an enterprise setup the focus might be set on certifying quality in terms of indicating faults rather than blaming a particular party for potential attacks along the FL process. 


In the following section, we introduce the proposed \proj, which enables accountability in FL.
\section{Overview of our \proj}
\label{sec:overview}



\begin{figure*} 
    \centering
  \subfloat[Processes. First four processes are subject to accountability to enable auditing through the use of FL FactSheets\label{fig:procceses}]{%
       \includegraphics[width=0.8\linewidth]{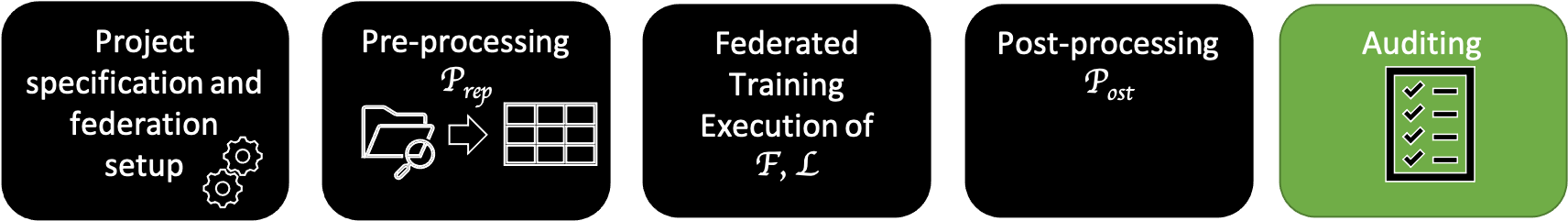}}
    \hfill \\
  \subfloat[Framework Architecture\label{fig:architecture}]{%
        \includegraphics[width=0.8\linewidth]{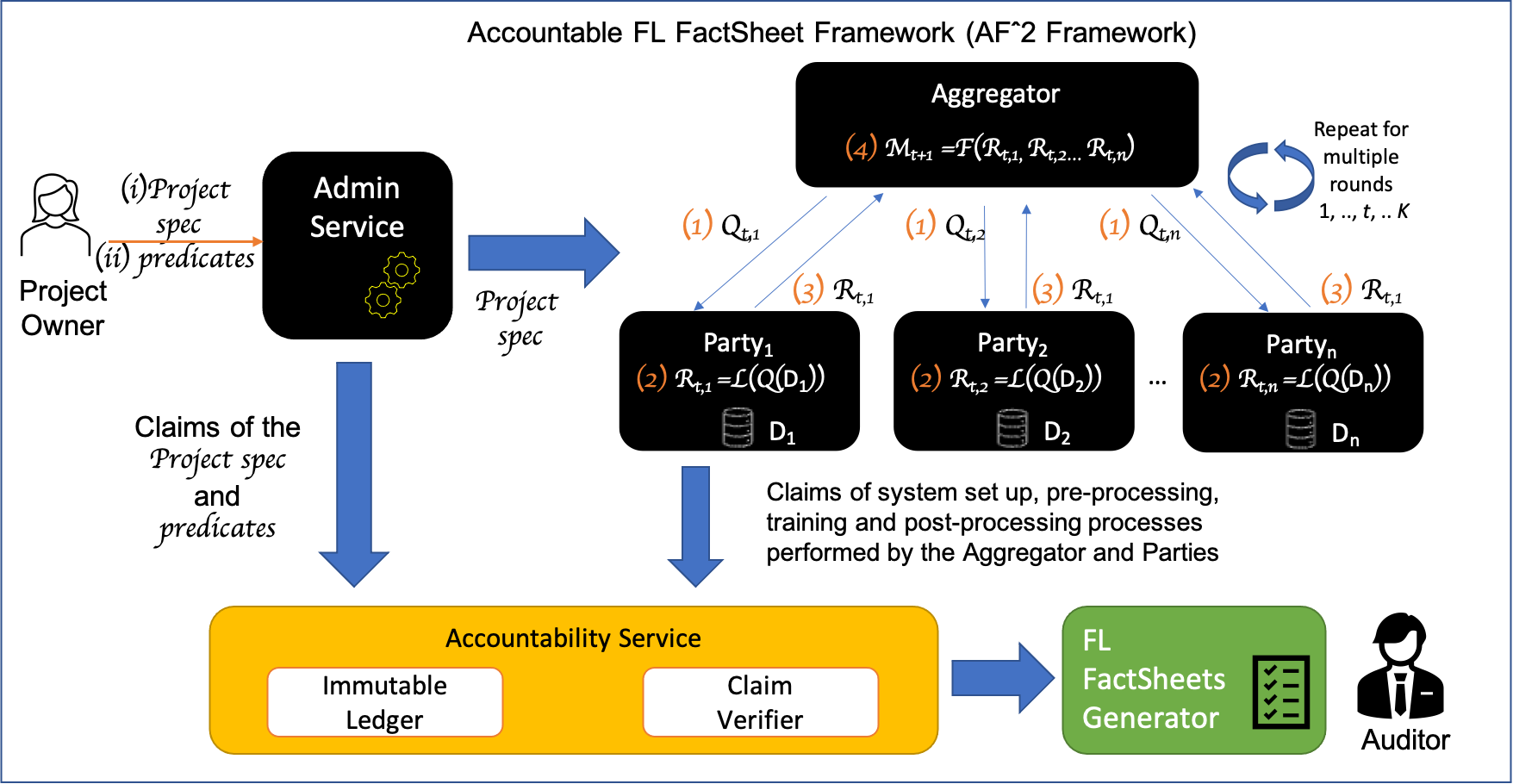}}
    \hfill 
  \caption{Overview of the \proj}
  \label{fig:framework}
\end{figure*}

The goal of our \proj is to bring accountability into the complete life cycle of the FL process shown in Figure \ref{fig:procceses}, allowing for reproducible results and detection of any misbehaviors during the training process, that we make available to all involved actors through corresponding factsheets.

We define two roles: a \textit{project owner} and an \textit{auditor}\footnote{We acknowledge the existence of numerous roles in ML and FL (cf. e.g. \cite{raji2020closing}), but focus on these particular ones in our current research}.
The \textit{project owner} defines the federation and prescribes the model, hyperparameters, pre-processing and any post-processing algorithms to be used.
It is in the project owner's best interest to ensure that the federation produces high-quality models and that the process 
is transparent, with sufficient information logged for auditing purposes or in case of any other problems.
Additionally, the project owner defines the \textit{predicates} to be verified by \proj.
For example, a predicate may verify that all parties' replies have been equally added into the FL process, or that a fairness post-processing methodology was applied.

The \textit{auditor} is an entity that can audit the FL process by
verifying how the model was trained.
The auditor needs evidence to determine if the process followed had any potential issues. 
For example, a project owner may be embodied by a consulting firm, a consortium lead or by a group of people representing all parties,
while an auditor may be internal or external.

Our \proj provides a way for the project owner to collect enough information to ensure an auditor can verify the complete FL life cycle.
Moreover, it provides the auditor a single \textit{sheet} where they can see the process carried out and determine whether any of the steps followed does not comply with pre-specified desired behaviors.

Figure \ref{fig:architecture} presents the architecture of the \proj,
where besides the aggregator and parties, an \textit{Admin Service},
an \textit{Accountability Service} and an \textit{FL FactSheet Generator}
are introduced.
The \textit{Admin Service} is the interface for the project owner to provide the specification of the project. 
The \textit{Accountability Service} ensures that the claims stored by other entities in the framework are immutable and can be traced back to a particular entity (i.e.~non-repudiability).
It also ensures that the project owner's predicates describing the expected behavior of the project
are evaluated on logged claims through the \textit{Claim Verifier}.
The \textit{FL FactSheet Generator} takes this verified information with all claims and
creates a view for the auditor to identify at a glance whether any parts of the
prescribed process were not followed.

We now describe the first four FL processes presented in Figure \ref{fig:procceses}
and the claims stored in the \textit{Accountability Service} to enable accountability.

\textit{Process 1: project specification and federation set up.} The project owner defines the \textit{project specification} for the federation.
This may include the hyperparameters, fusion algorithm, local training handler, expected data format and any post-processing steps performed on the model. 
All this information is received by the \textit{Admin Service}, which distributes the project specification to the Aggregator and Parties.
These entities use the project specs to set up the federation.
For accountability, the Admin Service, Aggregator and Parties create claims of the project specs they used and record them in the Accountability Service.
The Claim Verifier receives predicates to verify stored claims corresponding to the ones provided by the project owner.
In this way, the auditor can at a later stage verify the system set up was performed according to the owner's project specifications, not alternative ones.

\textit{Process 2: Pre-processing.} As part of the project specs, the data owner defines the data format, and each party is responsible for ensuring its training data complies with this expected format. For example, all columns should be in the same order for all parties and they should be normalized using the same algorithm. This process may be error prone and is critical for the success of the federation. Furthermore, any failure in this step may result in poor model performance.
For these reasons, each party needs to add claims for its data pre-processing to the Accountability Service.
Note that each party may carry out a slightly different pre-processing algorithm to account for differences in how their raw data is originally collected or stored.
For this reason, each party should claim information about the concrete steps that where followed, and if available, information about the provenance of the raw data.

In some cases, the Aggregator may also have a hold-out set of data that is used to evaluate the model performance during the training process.
If this is the case, the Aggregator is also required to transform its data to comply with the required format. Similarly to the parties, claims about this process are also stored for accountability purposes.


\textit{Process 3: Training.} During this process the fusion algorithm and local training process are run using the specified hyperparameters by the aggregator and parties.
As a result, a single final model can be obtained.
The aggregator and parties need to store claims of the critical steps in the fusion algorithms $\mathcal{F}$ and local training process $\mathcal{L}$, respectively.
These include information about the hyperparameters used, information received from other entities, and the concrete algorithm run.
The Claim Verifier Module receives predicates to verify that the critical steps followed pre-specified guidelines.
For instance, a predicate may verify the fair inclusion of all parties in the training process.
In the following section, we explain in detail the claims that need to be recorded to fully verify and replicate the training process.

\textit{Process 4: Post-processing (optional).} This process is optional and can have multiple purposes, for example, in certain federations the aggregator may be entrusted to add differentially private noise to the model to prevent inference threats. In other cases, the aggregator may run techniques to mitigate bias and provide a fair model. This process modifies the final model and related claims need to be logged.

After all the processes above conclude, the Claim Verifier uses the provided predicates to evaluate the process, and the FL FactSheets Service generates a FactSheet with the Verifier's results and all recorded claims. 
The auditor can quickly and reliably determine if there was an error during the FL process, and perform further in-depth verifications if needed.

In the following, we demonstrate how \proj operates in detail.


\begin{algorithm}
\SetAlgoLined
\KwActors{Admin Service $\mathcal{O}$; Accountability Service $\mathcal{S}$; Aggregator $\mathcal{A}$; Parties $P_i$ for $i=1,2,\dots,n$.}
\KwRoutines{Data preprocessing $\mathcal{P}_{re}$; Local training $\mathcal{L}(\cdot, \cdot)$; fusion $\mathcal{F}(\cdot)$; hash function $\mathcal{H}$; entity adds claim to $\mathcal{S}$ using its private key to sign it \textit{add\_claim(info$_1$,\dots,info$_l$)}}
\KwHyperparameter{Number of iterations $K$; number of parties $n$; 
local training hyperparameters $\eta_l$; global fusion hyperparameters $\eta_g$.}
\KwInputData{Each party $P_i$'s local data $D_i$ for $i=1,2,\dots,n$.}
 \tcp{Project spec and FL set up:}\label{ln:setup:init}
 $\mathcal{O}$, $\mathcal{S}$ and all $P_i$ generate crypto keys \label{ln:keysetup}\\
 $\mathcal{O}$ receives \textit{spec} = $\langle \mathcal{P}_{re}$, $\mathcal{L}(\cdot, \cdot)$, $\mathcal{F}(\cdot)$,
 $\mathcal{H}$, $K$, $n$, $\eta_l$, $\eta_g \rangle$ \\ 
 $\mathcal{O}$  \cprepro{\textit{add\_claim(spec)}} \tcp*[r]{{Owner's project spec is signed and logged}} \label{ln:spec:o}
 $\mathcal{O}$ sends \textit{spec} to $\mathcal{A}$ and $P_i$ for $i=1,2,\dots,n$.\\
 $\mathcal{A}$ initializes its system using \textit{spec} \\
 $\mathcal{A}$ \cprojspec{\textit{add\_claim(spec)}} \tcp*[r]{{Aggregator signs and logs the spec. it uses }} \label{ln:spec:A}
 \tcp{{All parties pre-process their data: }}
 \ForParallel{$i=1,2,\dots,n$} {
 $P_i$ initializes its system using \textit{spec} and \cprojspec{\textit{add\_claim(spec)}} \tcp*[r]{{Each party signs and logs the spec it uses}} \label{ln:setup:end} \label{ln:spec:P}
 $P_i$ \cdataprov{\textit{add\_claim(}$\mathcal{H}(D_i)$, \textit{provenance}($D_i$)\textit{)}}  \tcp*[r]{{Each party $P_i$ hashes its data, signs its provenance information and logs it}}
 $P_i$ pre-process its data set $D_i$ = $\mathcal{P}_{re}(D_i))$ and \cprepro{\textit{add\_claim(}$\mathcal{H}(D_i)$,$\mathcal{P}_{re}$\textit{)}} \tcp*[r]{{ Each party $P_i$ adds a claim stating the pre-processing routine.}} \label{ln:preprocess:data}
 }
 \tcp{{Training process:}}
\For{$t=1,2,\dots,K$}{
\ForParallel{$i=1,2,\dots,n$} {
  \tcp{{Aggregator generates queries:}} \label{ref:queries}
  $\mathcal{A}$ generates new queries $\mathcal{Q}_{t,i}$ using $M_t$ where $M_t$ is the initial model\\
  $\mathcal{A}$ \ctrain{\textit{add\_claim(}$Q_{t,i}$\textit{)}} \tcp*[r]{{Aggregator claims the query for round $t$ sent to party $i$, $Q_t,i$ for all parties, }}  \label{ref:queries:A:claim}
  $\mathcal{A}$ sends $Q_{t,i}$ to $P_i$\\
  \tcp{{Parties train:}} 
  $P_i$ \ctrain{\textit{add\_claim(}$Q_{t,i}$\textit{)}} \tcp*[r]{{Each party $P_i$ claims the query it received from the aggregator }}  \label{ref:queries:P:claim} 
  $P_i$ runs  $R_{t,i} \gets \mathcal{L}(\mathcal{Q}(\eta_l, D_i)$) \tcp*[r]{{Party $P_i$ runs the local training process $\mathbf{L}$ using the received query $\mathcal{Q}$, hyperparams $\eta$ and its local dataset $D_i$ }} 
  $P_i$ computes local model update $R_{t,j} \gets \mathcal{L}_i(\mathcal{Q}_{t,i}(D_i)$ \\
  $P_i$ \ctrain{\textit{add\_claim(}$\mathcal{L}$, $\eta_l$, $\mathcal{H}(D_i)$ $R_{t,j}$\textit{)}}\label{ln:party:training:claim} \\
  $P_i$ sends $R_{t,j}$ to $\mathcal{A}$ \\
}
$\mathcal{A}$ waits for quorum \\ 
\textbf{if quorum:  } \\
$~~~\mathcal{A}$ determines parties in $Q \subseteq \{P_1,\dots, P_n\}$ \tcp*[r]{{Aggregator determines what parties replied }}
$~~~\mathcal{A}$ computes $M_{t+1} \gets \mathcal{F}(R_{t,1}, R_{t,2}, \dots, R_{t,n})$ \tcp*[r]{{New model $M_t+1$ is found by applying fusion algorithm $\mathcal{F}$ using received party replies }}
$~~~\mathcal{A}$ \ctrain{\textit{add\_claim(}$M_t$, $Q$,  $(R_{t,1}, R_{t,2}, \dots, R_{t,n})$\textit{)}} \tcp*[r]{{Aggregator claims the the information it used to run the fusion algorithm, including all party replies used in the process}} \label{ln:fusion:claim:A}
\textbf{else:} $\mathcal{A}$ \ctrain{\textit{add\_claim(No quorum)}} and stops \tcp*[r]{{Aggregator claims that there wasn't a quorum.}} 


}
\tcp{{Post-processing process: }}
$\mathcal{A}$ runs post-processing method (optional): 
 $M_f = \mathcal{P}_{post}(M_K)$ \\
 $\mathcal{A}$ \cpostpro{\textit{add\_claim(}$\mathcal{P}_{post}$, $M_f$\textit{)}} {\tcp*[r]{{Post-processing procedure is signed and added.}}}
 FL FactSheets are generated to present the predicates, their resulting evaluation and claims collected.
 \caption{\proj}
\end{algorithm}

\section{\proj Detailed Specification} 
\label{sec:factsheet-details}
In this section, we provide the specification of the system showing how the standard FL process needs to be modified to provide accountability. Our specification is general and can be applied to multiple predicates that are relevant for accountability during the different stages of the FL process. 
For that reason, we provide an in-depth look at how the FL process is modified to fulfil the objectives of the system in Algorithm~1, where we use as basis the notation presented in the background section.
Additionally, we provide a higher level overview in Table \ref{tab:predicates-claim-examples} where we demonstrate useful predicates and how they can be successfully evaluated using the information logged in Algorithm~1. 

\subsection{Undeniable and tamper-evident claims} 
To enable accountability, it is necessary to ensure that the claims added to the \textit{Accountability Service}, $\mathcal{S}$,
are undeniable and tamper-evident. For this purpose, during system set up, all entities in the system
need to obtain access to $\mathcal{S}$ and generate their cryptographic keys $\langle sk, pk \rangle$,
where $sk$ is the private key and $pk$ its public key.
The Accountability service $\mathcal{S}$ needs to know the public key of all other entities. This trust-establishing step is presented in line \ref{ln:keysetup} of Algorithm~1. \footnote{A trusted execution environments could also be used to enable parties and aggregators to attest their code ensuring the right process is followed.
If this is not available, we will see that the claims can still be verified and we would still be able to verify whether the aggregator and parties have a different view of the system}

With this key setup, we are now able to log claims in a distributed immutable ledger. A distributed ledger is a combination of computer science, cryptography, and economic mechanisms for the coordination of participants towards building consensus \cite{wattenhofer2016science} and “secure processing of transactions between untrustworthy parties in a decentralized system” \cite{kannengiesser2019does}, while maintaining a single point of truth \cite{wattenhofer2016science}.
We define a function \textit{add\_claim}($\mathcal{S}$, $sk$, (\textit{info}$_1$, \textit{info}$_2$, \dots))
that enables any entity of the system to store any list of information (\textit{info}$_1$, \textit{info}$_2$, \dots) in the Immutable Ledger of $\mathcal{S}$. This function also takes as input the private key $sk$, which is used
to sign the claim before sending it to $\mathcal{S}$. In this way, the origin of the claim is guaranteed to be from the owner of the key $sk$.
For simplicity in notation, whenever it is clear by the context, we use \textit{add\_claim}(\textit{info}$_1$, \textit{info}$_2$, \dots) without explicitly mentioning $\mathcal{S}$ and the private key.
Note that all claims sent to the Accountability service $\mathcal{S}$ by $\mathcal{O}$, $\mathcal{A}$ and all parties $P_i$
are signed with their private key, therefore $\mathcal{S}$ can easily verify that the messages are sent by these entities by checking the validity of the signature.
In this way, all claims are undeniable.

{
\begin{table}
  \centering
  \begin{tabular}{p{27mm}|p{55mm}|p{55mm}}
    \hline\noalign{\smallskip}
    & Example predicates to verify & Evidence to verify claims \\
    \hline
    
    {\vspace{0cm}\includegraphics[height=2.0cm]{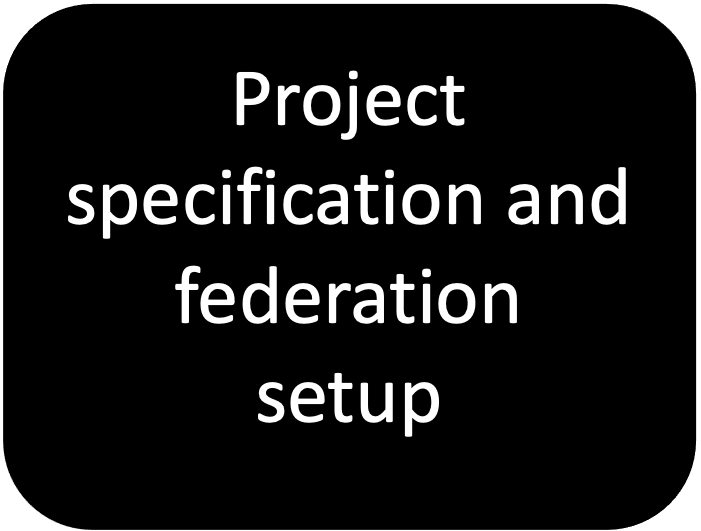}}
    & {Are the aggregator and parties using the correct algorithms? 
    \begin{itemize}[leftmargin=*]
        \item Regulated environments may require SMC such as fully homographic encryption
        \item Noisy or partially trusted environments require robust algorithms e.g., Krum
    \end{itemize}
}
 & {\begin{itemize}[leftmargin=*]
     \item \cprojspec{Line 6}: Spec used to load the aggregator’s source code including its hash
    \item \cprojspec{Line 8}: Spec/hash of used parties’ source code. 
 \end{itemize}} \\

    \noalign{\smallskip}
    \hline
    \noalign{\smallskip}


    
    \parbox[t]{35mm}{\vspace{10mm}\multirow{2}{*}{{\includegraphics[height=2.0cm]{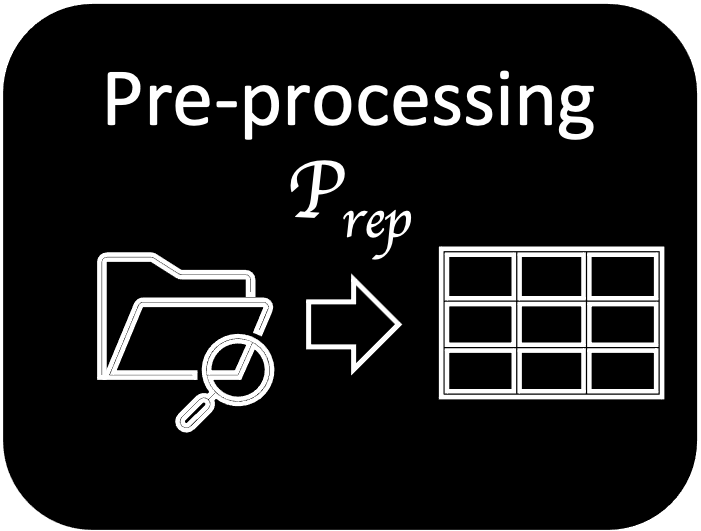}}}}
    & Did training data originate from a particular database?  
    What was the training data used? 
    & \cdataprov{Line 9}: Provenance of training data \\
    
    \cmidrule(lr){2-3}

    & {Was data pre-processed adequately? 
    \begin{itemize}[leftmargin=*]
        \item Was tabular data organized in the same order by all parties?
        \item Was a fairness pre-processing technique applied? 
    \end{itemize}} 
    & {\begin{itemize}[leftmargin=*]
        \item \cprepro{Line 3}: Spec defined by  project owner including data format expected and pre-processing techniques 
        \item \cprepro{Line 10}: Pre-processing process applied 
    \end{itemize}
} \\

    \noalign{\smallskip}
    \hline
    \noalign{\smallskip}

    \parbox[t]{35mm}{\vspace{18mm}\multirow{3}{*}{{\includegraphics[height=2.0cm]{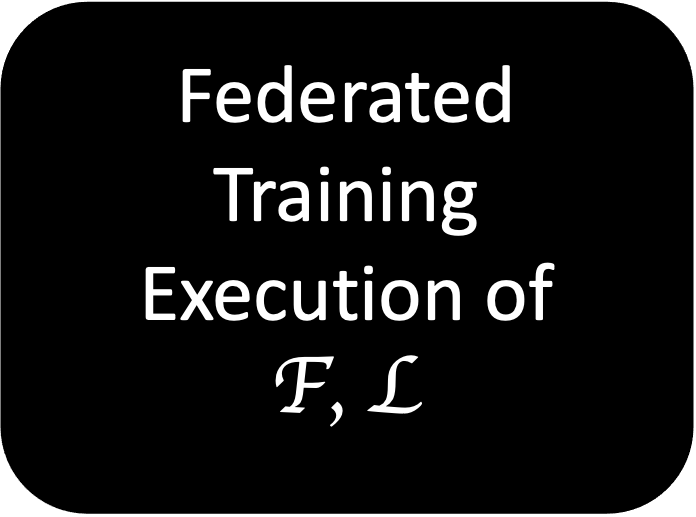}}}}
    & {Did the aggregator follow the correct fusion procedure?} 
    & {\begin{itemize}[leftmargin=*]
        \item \ctrain{Line 15}: Aggregator generated right query for parties 
        \item \ctrain{Lines 27/28}: Aggregator included all party replies as per expected algorithm 
    \end{itemize}}\\
    \cmidrule(lr){2-3}
    & {Did parties follow the correct local procedure? 
Are hyperparameters what they should be? E.g. If differential privacy is required, was the epsilon set up correctly? }
    & {\ctrain{{}Line 20, line 8}:, Hash of the data used for training, hyperparameters, local procedure} \\
    \cmidrule(lr){2-3}
    & {Do the parties and aggregator views of the process match?
Were the responses sent by the parties to the aggregator indeed included as part of the final model? Did the process \textit{fairly} include all parties to ensure data from all parties influences the shared model?}
    & \begin{itemize}[leftmargin=*]
        \item \ctrain{Lines 15, 17}: query claimed by the aggregator is the same received by the party
        \item \ctrain{Lines 21, 27}: model updates used by the aggregator are the ones sent by parties
    \end{itemize}\\

    \noalign{\smallskip}
    \hline
    \noalign{\smallskip}
    
    {\vspace{0cm}\includegraphics[height=2.0cm]{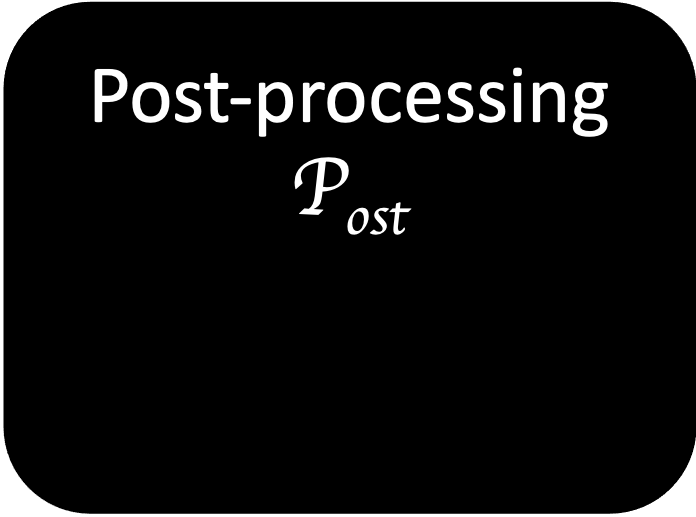}}
    & {Was the resulting model fine-tuned with dataset X? Was a fairness processing technique run to ensure model fairness?} 
    & {\begin{itemize}[leftmargin=*]
        \item \cpostpro{Line 3}: Spec defined by  project owner including post-processing technique(s)
        \item \cpostpro{Line 31}: post-processing information 
    \end{itemize}
} \\
    \hline
  \end{tabular}
    \caption{Sample predicates for common FL projects classified by process. The last column shows what information is used to evaluate the claims as per Algorithm~1.}
    \label{tab:predicates-claim-examples}
\end{table}
}
\subsection{Enabling predicate verification}


We now explain using a top-down approach how \proj works to enable accountability for
different predicates in the FL life-cycle.
Table \ref{tab:predicates-claim-examples} showcases different predicates for each of the processes highlighted in Figure \ref{fig:procceses}.
The first column shows the process, the second, useful predicates showcasing different scenarios where the information is relevant
and finally, the third column specifies what information is used to evaluate the predicates as specified in the modified FL process shown Algorithm~1.
This algorithm defines what concrete claims need to be logged by each entity in the system. 
It is important to note that the predicate and its evaluation will be included in the FactSheet.






 \begin{enumerate}
    \item Process 1: 
    The first sample predicate relates to the system's set up: \textit{are the aggregator and parties using the correct algorithms? }
    This is relevant in highly regulated environments that for example require fully homomorphic encryption or other types of FL extensions such as
    differential privacy or the application of a robust algorithm. Even if the FL collaboration does not require stringent security or privacy additions, verifying this predicate enables reproducibility and helps determine correctness.
    To verify this predicate, the FL process is modified to ensure that the \textit{project owner} claims the correct configuration.
    Additionally, the aggregator and parties need to claim the source code that they are using.
    The Accountability Service verifies the claim and stores the result in the FactSheet. 
    
    \item Process 2: 
    Correctly performing pre-processing is key for the FL process to run smoothly. Let's analyze a few important sample predicates.
    Consider the following predicates \textit{What data was used?} \textit{Was the training data coming from a particular database?} 
    This is relevant, if for example, we only want to train the ML model based on patients' data who have granted consent for a given study.
    To this end, the parties need to log information on the data provenance of the samples used.
    
    Another predicate of interest is \textit{Was data pre-processed adequately by each party?}
    In FL this represent a extremely relevant step that may be error prone. With each party maintaining their data locally, there is lack of visibility on the correctness of this process. For example, consider a healthcare use case where a party may use a dataset where the first column of a table is \textit{age}, while a second party may have the \textit{weight} of the patient. This clearly would lead to an incorrect, and even dangerous, resulting model. By evaluating the predicate, using the logged information (see third column), it is simple to determine if there is an error.
    In another example, the project owner or auditor may be interested in knowing if an ML fairness pre-processing technique was followed by each party.
    After the verification of these claims, the FactSheet is also updated with the results. 

    
    \item Process 3:
    In this process we evaluate the correctness and procedures followed during the training by all entities in the system.
    Sample predicates include \textit{Did the aggregator follow the correct fusion procedure?}
    \textit{Did parties follow the correct local procedure?} \textit{Are the hyperparameters what they should be?} for example, if differential privacy is required, was the epsilon value used the same as the one specified by the project owner? 
    And \textit{Do the parties and aggregator views of the process match?}
    Using the information described in the Table \ref{tab:predicates-claim-examples}, the Accountability Service can use the evidence submitted by each entity to evaluate the state of the system during multiple rounds.
    
    Other predicates may evaluate the multi-round behavior of the system, for example verifying that all parties have been included in the training process rounds the same number of times. This may be relevant if different parties have information that is unique and needs to be represented to avoid, for example, the creation of race-biased models that result from the under inclusion of some parties (with data representing various ethnic groups).

    \item Process 4:
    Finally, in some cases a post-processing step is needed. Example claims include 
    \textit{Was the resulting model fine-tuned with dataset X?}
    This predicate would determine if for example, the final model is personalized for a particular city or patient.
    And predicate \textit{Was a fairness processing technique run to ensure model fairness? } is another example where an auditor or project owner can understand what type of model was generated and also reproduce the process if needed.

\end{enumerate}


In what follows, we present our FL FactSheet implementation.

\section{Prototype and Use Case}
\label{sec:prototype}

\subsection{Prototype implementation}

\begin{figure*}[h]
    \centering
    \includegraphics[scale=.4]{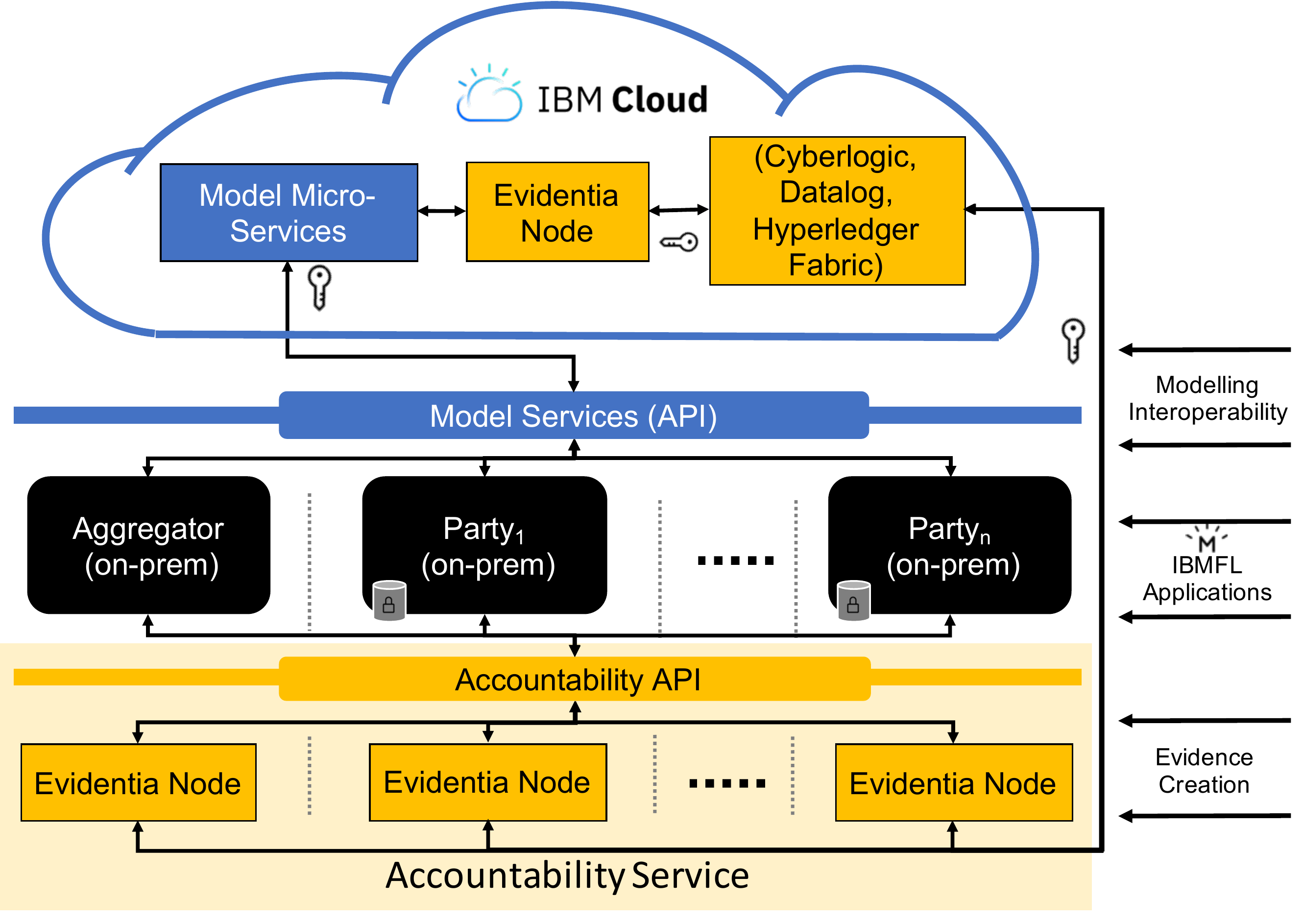}
    \caption{\proj prototype architecture (high-level overview)}
    \label{fig:prototype}
\end{figure*}

The current implementation of the \proj is based on the technologies illustrated in Figure~\ref{fig:prototype}. 
The prototype is an instantiation of the framework architecture (cf. Figure~\ref{fig:architecture}).

\textbf{Accountability Service Setup:}
To implement the Accountability Service, we introduce \textit{Evidentia Nodes} which use \textit{Hyperledger Fabric}{\footnote{\url{https://www.hyperledger.org/use/fabric}}} in the backend and the \textit{Accountability API}.

Evidentia{\footnote{\url{https://git.fortiss.org/evidentia}}} is built on top of Datalog~\cite{greco2015datalog} for the implementation of distributed logic programs as well as of Hyperledger Fabric~\cite{androulaki2018hyperledger,cachin2016architecture}, a Distributed Ledger Technology, ensuring tamper-proof logging. Additionally, Hyperledger Fabric is responsible for ID management. Consequently, Evidentia allows auditors to infer from the logs which party is responsible for which evidence and which parts of the formalised specifications.



To implement the aggregator and party functionality, we \textit{extended} the IBM FL framework \cite{ibmfl} 
with the Musketeer plugin that enables a \textit{RabbitRabbit MQ} connection.
\textit{Rabbit MQ} enables outbound-only network connections using queues where each entity in the system (aggregator and parties) can communicate by leaving messages in the queue.
We opted for this connection to ensure participants do not need to open ports in their local machines 
(obtaining such permissions is cumbersome and frequently requires approval from the Security Unit in enterprises in highly regulated domains).
The queues for the Rabbit MQ connection are 
Outbound-only network connections are initiated from participants to the cloud services, via the \textit{Model Services API}. This enhances the security features of the prototype.


%
\textbf{Federated Learning Setup:}
To implement the aggregator and party functionality, we \textit{extended} the IBM FL framework \cite{ibmfl}
by adding the \textit{Model Services API} and cloud-based \textit{Model Micro-Services}. 
The role of the \textit{Model Services API} is to facilitate communications between the aggregator and parties via the cloud using a queue-based service. 
All exchanged information is routed using the \textit{Model Micro-Services}, ensuring that the appropriate entities receive the required content. 

The evidence for each interaction between parties and the aggregator is stored in \textit{Evidentia Nodes} using the \textit{Accountability API}.
\textit{Evidentia Nodes} run within a Kubernetes cluster on the cloud and receive evidence from the micro-services, dispatching this evidence to the \textit{Hyperledger}. 
We also extended the IBM FL framework to enable the aggregator and parties to store evidence claims via the \textit{Accountability API} and dedicated \textit{Evidentia Nodes}.

Finally, exchanged models are stored in an encrypted object store, with references to these models and general interactions stored in a relational database. This enables a retrospective review of the FL process, providing a full lineage for the final model.

\subsection{A Use Case of Accountable FL based on FactSheets}

\textbf{Case background:} As an exemplary application domain of accountable FL, we have chosen an online citizen participation as a governmental use case that requires high requirements on compliance with ethics guidelines and principles. Online  citizen  participation  can  be  described  as  a  form of  participation  that  is  based  on  the  usage  of  information and  communication  technology in  societal  democratic and consultative  processes  focused  on  citizens.

The data used for the prototype was collected during several citizen participation sessions in Germany. The dataset is a collection of multiple texts submitted by citizens written in German, each describing ideas which include the description of their problems, concerns and suggestions. The texts may also contain other information like title and category. Multiple cities engaged in FL to create a neural network model that facilitates the classification of each submitted idea to expedite routing of the received ideas. Because the training set may contain private data and there is no legal basis for sharing such data, FL is key to enable this use case.  


\begin{figure*}[h]
    \centering
    \includegraphics[scale=.17]{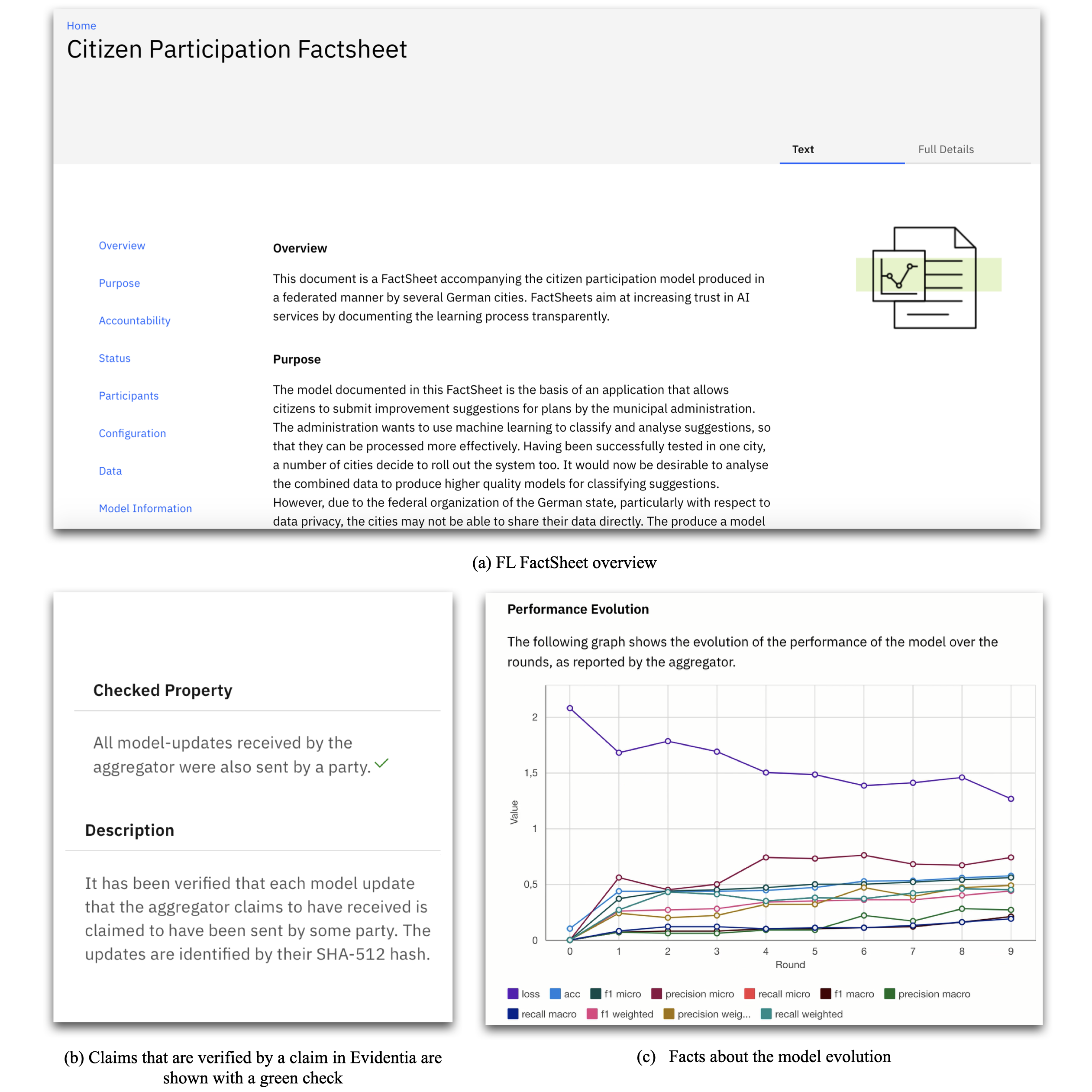}
    \caption{Screenshots of the \proj generated FL FactSheet}
    \label{fig:factsheet}
\end{figure*}

\noindent\textbf{Factsheet sketch:}
For narrative purposes, Figure \ref{fig:factsheet} presents an overview of \proj generated FL FactSheets with a human-readable and easily comprehendible representation of claims and corresponding evidence.
A complete FL FactSheet can be seen here: \url{https://ibm.biz/Bdf55q}.
After an overview of the content, a table is populated with the \textit{predicates} (e.g. checked properties) and description as shown in Figure \ref{fig:factsheet}b. The claims that are \textit{verified} by a predicate in Evidentia are shown with a green check ensuring that a party or an auditor can quickly know they are correct, and a drill-down for supporting evidence can be seen. Model performance is another point of information that is digested from the available predicates. In the sketch, the system includes multiple metrics of the model as depicted in Figure \ref{fig:factsheet}c.
With all this information, it is finally possible to decide if there was an unexpected or erroneous situation and what model to select (in case multiple projects have been launched).

\subsection{Exemplary claims and evidence}
{\footnotesize
\begin{table}
  \centering
  \begin{tabular}{p{25mm}|p{35mm}|p{42mm}|p{35mm}}
    \hline\noalign{\smallskip}
    & Concrete examples for predicates  & Concrete examples of evidence to verify claims & Corresponding log entry\\
    \hline
    {\vspace{0cm}\includegraphics[height=2.0cm]{figures/1_proj_spec.png}}
    & {\begin{itemize}[leftmargin=*]
        \item Type of algorithm selected for aggregation(e.g. for more robustness)
        \item ML Model used for training
        \item Global hyperparameters(number of parties, number of global rounds, stopping criteria...)
        \item Local hyperparameters(learning rate, training epochs...)
        
    \end{itemize}}
 & { \begin{itemize}[leftmargin=*]
     \item ``KrumFusionHandler"
     \item ``Keras-CNN" 
     \item Maximum Timeout: 600 seconds, Number of parties: 5, Number of rounds: 10, Termination accuracy: 90\%, Learning Rate: 0.1, Number of epochs: 1
     
 \end{itemize}}
 & {\begin{itemize}[leftmargin=*,align=left]
     \item \texttt{fusion\_algorithm (``KrumFusionHandler'')}
     \item \texttt{model\_name(``Keras-CNN'')}
     \item \texttt{max\_timeout(600), parties(5), rounds(10), termination\_accuracy(0.9)}
    
 \end{itemize}}\\
    
    \noalign{\smallskip}
    \hline
    \noalign{\smallskip}

%
%

    {\vspace{0cm}\includegraphics[height=2.0cm]{figures/2_pre_proc.png}}
    & {
    \begin{itemize}[leftmargin=*]
        \item Selected data handler
        \item Data provenance: Meta Data (length of training/testing set, label names...)
        \item Data provenance: Hash of the training set
        \item Data provenance: Hash of the testing set
    \end{itemize}}
    & {
    
    \begin{itemize}[leftmargin=*]
     \item ``CivitasKerasDataHandler"
     \item Party 0 has training data of size 703, provenance hash value: ``ca6f39a8f01f9ba..."
      \item Aggregator has test data of size 391", provenance has value: ``672b8b2258538..."
 \end{itemize}}
    &{\texttt{
    \begin{itemize}[leftmargin=*,align=left]
     \item data\_handler (``KrumFusionHandle'')
      \item party0~attests \texttt{\small training\_data\_size(703)}
      \item Aggregator~attests \texttt{\small test\_data\_size(391)}
learning\_rate(0.1), epochs(1)
 \end{itemize}
    }}\\

    \noalign{\smallskip}
    \hline
    \noalign{\smallskip}

    {\vspace{0cm}\includegraphics[height=2.0cm]{figures/3_FL_train.png}}
    & { \begin{itemize}[leftmargin=*]
        \item Hash of the model updates
        \item Selected models for aggregation
        \item Evaluation metrics(accuracy, F1 score, recall, precision...)
    \end{itemize}}
    & { \begin{itemize}[leftmargin=*]
        \item Aggregator sent model with Hash ``a68b0e5989a..." at round 3
        \item Aggregator selected the model update number 4 at round 5
        \item Metrics at round 2: Accuracy 43,7\%, F1-Score 44\%
    \end{itemize}}
    & { 
    \begin{itemize}[leftmargin=*]
        \item Aggregator attests \texttt{\small sent\_global\_model(3, "a68b0e5989a…")}
        \item Aggregator attests \texttt{\small selected\_model\_update(5, 4)}
        \item Aggregator attests \texttt{\small evaluation\_results(2, '{ "acc": 0.43734, "f1 micro": 0.44, ...}')}
    \end{itemize}}\\
    
    \hline
  \end{tabular}
    \caption{Table with examples for claims and evidence for our prototype implementation}
    \label{tab:claim-evidance-examples}
\end{table}
}

We briefly explain how we ensure verifiable, undeniable, auditable and tamper-evident of the claims. Table \ref{tab:claim-evidance-examples} shows some concrete examples of information logged in our prototype.

\begin{itemize}
    \item \textbf{Verifiable:} we use different predicates. For instance, information is provided regarding the project specification in terms of which type of algorithm is selected for aggregation with respect to e.g. increased robustness. A piece of evidence is provided as well, in that case ``KrumFusionHandler" is the selected type of algorithm. 
    
    \item \textbf{Undeniable:} we automatically check, if all participants share the same required knowledge. For instance, models updates are acknowledged by both sending aggregator and a receiving party based on a unique hash. If someone does not acknowledge this information, the FactSheet will contain this information and will apply corresponding rules to the overall claim of the process.

    \item \textbf{Auditable:} we provide detailed information that can be used in reproducing the process. For instance, we generate evaluation metrics about the performance of potentially applicable learning algorithms and support the argument of choosing one of the latter.
    
    \item \textbf{Tamper-evident:} we store  metadata as well as hashes of the data. For instance, we store information about training data size and the corresponding hash can be used to resolve potential issues if the data is altered or different data is provided as evidence.

\end{itemize}
\section{Conclusions}
FL is a promising approach towards privacy-preserving and ethical AI. In regulated environments, however, its adoption may be hindered due to the potential lack of accountability. To the best of our knowledge, our \proj is the first to address this challenge, allowing FL processes to generate verifiable, undeniable, certifiable and tamper-evident claims that guarantee for 
indicating faults in the distributed FL process that may have occurred when one of the entities in the system deviates from the protocols or there are byzantine errors, for reproducing the FL process, and for allowing for better understanding of the ML models generated, their biases and when they can be applied. The \proj supports the automatic verification of predicates based on the collected evidence along the claims, ensuring that non-compliance is easily detected.

We hope that the \proj and its implementation, lays the groundwork for a wider adoption of FL through accountability, reproducibility and overall increased trustworthiness. In particular, we emphasize the importance of the ability of the framework to document the process we identified and presented, and thus to comply with requirements for process verification (as e.g. proposed in the European Commission draft \cite{whitepaperEUAI2020}).

While we have shown an initial effort on the operationalization of accountability and reproducibility for FL based on exemplary claims and corresponding evidence, we believe that an interesting direction for future research includes exploring holistic claims derived from legislative or contractual rules, guidelines etc., as well as introducing automated auditing through trusted execution environments. Future research could also address the greater number of roles involved in the ML and FL process (cf. e.g. \cite{raji2020closing}), corresponding claims and coping with the trade-off between privacy and required transparency. Further research endeavours might involve secure and privacy-preserving reproduceability of the FL process steps (e.g. through cryptographic techniques \cite{walfish2015verifying}).

\clearpage

\bibliographystyle{unsrt}
\bibliography{references.bib}

\begin{thebibliography}{10}

\bibitem{jobin_global_2019}
Anna Jobin, Marcello Ienca, and Effy Vayena.
\newblock The global landscape of {AI} ethics guidelines.
\newblock {\em Nature Machine Intelligence}, 1(9):389--399, 2019.
\newblock Publisher: Nature Publishing Group.

\bibitem{leslie_david_2019_3240529}
David Leslie.
\newblock {Understanding artificial intelligence ethics and safety: A guide for
  the responsible design and implementation of AI systems in the public
  sector}, June 2019.

\bibitem{whittlestone_role_2019}
Jess Whittlestone, Rune Nyrup, Anna Alexandrova, and Stephen Cave.
\newblock The role and limits of principles in {AI} ethics: towards a focus on
  tensions.
\newblock In {\em Proceedings of the 2019 {AAAI}/{ACM} Conference on {AI},
  Ethics, and Society}, pages 195--200, 2019.

\bibitem{mittelstadt_principles_2019}
Brent Mittelstadt.
\newblock Principles alone cannot guarantee ethical {AI}.
\newblock {\em Nature Machine Intelligence}, 1:501--507, 2019.

\bibitem{wilson2021building}
Christo Wilson, Avijit Ghosh, Shan Jiang, Alan Mislove, Lewis Baker, Janelle
  Szary, Kelly Trindel, and Frida Polli.
\newblock Building and auditing fair algorithms: A case study in candidate
  screening.
\newblock In {\em Proceedings of the 2021 ACM Conference on Fairness,
  Accountability, and Transparency}, pages 666--677, 2021.

\bibitem{fedavgMahan}
Brendan McMahan, Eider Moore, Daniel Ramage, Seth Hampson, and Blaise~Aguera
  y~Arcas.
\newblock Communication-efficient learning of deep networks from decentralized
  data.
\newblock In {\em Artificial Intelligence and Statistics}, pages 1273--1282.
  PMLR, 2017.

\bibitem{apple}
Matthias Paulik, Matt Seigel, Henry Mason, Dominic Telaar, Joris Kluivers,
  Rogier van Dalen, Chi~Wai Lau, Luke Carlson, Filip Granqvist, Chris
  Vandevelde, et~al.
\newblock Federated evaluation and tuning for on-device personalization: System
  design \& applications.
\newblock {\em arXiv preprint arXiv:2102.08503}, 2021.

\bibitem{google}
{“Federated learning: Collaborative machine learning without centralized
  training data"}.
\newblock
  \url{https://ai.googleblog.com/2017/04/federated-learning-collaborative.html},
  2017.
\newblock [Online; accessed 29-Oct-2021].

\bibitem{webank}
{“Utilization of fate in risk management of credit in small and micro
  enterprises"}.
\newblock
  \url{https://www.fedai.org/cases/utilization-of-fate-in-risk-management-of-credit-in-small-and-micro-enterprises/},
  2019.
\newblock [Online; accessed 29-Oct-2021].

\bibitem{rieke2020future}
Nicola Rieke, Jonny Hancox, Wenqi Li, Fausto Milletari, Holger~R Roth, Shadi
  Albarqouni, Spyridon Bakas, Mathieu~N Galtier, Bennett~A Landman, Klaus
  Maier-Hein, et~al.
\newblock The future of digital health with federated learning.
\newblock {\em NPJ digital medicine}, 3(1):1--7, 2020.

\bibitem{kacianka2021designing}
Severin Kacianka and Alexander Pretschner.
\newblock Designing accountable systems.
\newblock In {\em Proceedings of the 2021 ACM Conference on Fairness,
  Accountability, and Transparency}, pages 424--437, 2021.

\bibitem{arnold2019factsheets}
Matthew Arnold, Rachel~K.E. Bellamy, Michael Hind, Stephanie Houde, Sameep
  Mehta, Aleksandra Mojsilović, Ravi Nair, Karthikeyan~Natesan Ramamurthy,
  Darrell Reimer, Alexandra Olteanu, David Piorkowski, Jason Tsay, and Kush~R.
  Varshney.
\newblock Factsheets: Increasing trust in ai services through supplier’s
  declarations of conformity.
\newblock {\em IBM Journal of Research and Development}, 63(4/5):6:1--6:13,
  2019.

\bibitem{kairouz2021advances}
Peter Kairouz, H.~Brendan McMahan, Brendan Avent, Aur{\'{e}}lien Bellet, Mehdi
  Bennis, Arjun~Nitin Bhagoji, Kallista~A. Bonawitz, Zachary Charles, Graham
  Cormode, Rachel Cummings, Rafael G.~L. D'Oliveira, Hubert Eichner, Salim~El
  Rouayheb, David Evans, Josh Gardner, Zachary Garrett, Adri{\`{a}}
  Gasc{\'{o}}n, Badih Ghazi, Phillip~B. Gibbons, Marco Gruteser, Za{\"{\i}}d
  Harchaoui, Chaoyang He, Lie He, Zhouyuan Huo, Ben Hutchinson, Justin Hsu,
  Martin Jaggi, Tara Javidi, Gauri Joshi, Mikhail Khodak, Jakub
  Kone{\v{c}}n{\'y}, Aleksandra Korolova, Farinaz Koushanfar, Sanmi Koyejo,
  Tancr{\`{e}}de Lepoint, Yang Liu, Prateek Mittal, Mehryar Mohri, Richard
  Nock, Ayfer {\"{O}}zg{\"{u}}r, Rasmus Pagh, Hang Qi, Daniel Ramage, Ramesh
  Raskar, Mariana Raykova, Dawn Song, Weikang Song, Sebastian~U. Stich, Ziteng
  Sun, Ananda~Theertha Suresh, Florian Tram{\`{e}}r, Praneeth Vepakomma, Jianyu
  Wang, Li~Xiong, Zheng Xu, Qiang Yang, Felix~X. Yu, Han Yu, and Sen Zhao.
\newblock Advances and open problems in federated learning.
\newblock {\em Found. Trends Mach. Learn.}, 14(1-2):1--210, 2021.

\bibitem{fedavg}
Keith Bonawitz, Vladimir Ivanov, Ben Kreuter, Antonio Marcedone, H~Brendan
  McMahan, Sarvar Patel, Daniel Ramage, Aaron Segal, and Karn Seth.
\newblock Practical secure aggregation for privacy-preserving machine learning.
\newblock In {\em Proceedings of the 2017 ACM SIGSAC Conference on Computer and
  Communications Security}, pages 1175--1191. ACM, ACM, 2017.

\bibitem{pfmn}
Mikhail Yurochkin, Mayank Agarwal, Soumya Ghosh, Kristjan Greenewald, Nghia
  Hoang, and Yasaman Khazaeni.
\newblock Bayesian nonparametric federated learning of neural networks.
\newblock In {\em International Conference on Machine Learning}, pages
  7252--7261. PMLR, 2019.

\bibitem{xgboostFLibm}
Yuya~Jeremy Ong, Yi~Zhou, Nathalie Baracaldo, and Heiko Ludwig.
\newblock Adaptive histogram-based gradient boosted trees for federated
  learning, 2020.

\bibitem{zeno}
Cong Xie, Sanmi Koyejo, and Indranil Gupta.
\newblock Zeno: Distributed stochastic gradient descent with suspicion-based
  fault-tolerance.
\newblock In {\em International Conference on Machine Learning}, pages
  6893--6901. PMLR, 2019.

\bibitem{tifl}
Zheng Chai, Ahsan Ali, Syed Zawad, Stacey Truex, Ali Anwar, Nathalie Baracaldo,
  Yi~Zhou, Heiko Ludwig, Feng Yan, and Yue Cheng.
\newblock Tifl: A tier-based federated learning system.
\newblock In {\em Proceedings of the 29th International Symposium on
  High-Performance Parallel and Distributed Computing}, New York, NY, USA,
  2020. Association for Computing Machinery.

\bibitem{hybridone}
Stacey Truex, Nathalie Baracaldo, Ali Anwar, Thomas Steinke, Heiko Ludwig, Rui
  Zhang, and Yi~Zhou.
\newblock A hybrid approach to privacy-preserving federated learning.
\newblock In {\em Proceedings of the 12th ACM Workshop on Artificial
  Intelligence and Security}, pages 1--11, 2019.

\bibitem{hybridalpha}
Runhua Xu, Nathalie Baracaldo, Yi~Zhou, Ali Anwar, and Heiko Ludwig.
\newblock Hybridalpha: An efficient approach for privacy-preserving federated
  learning.
\newblock In {\em Proceedings of the 12th ACM Workshop on Artificial
  Intelligence and Security}, pages 13--23, 2019.

\bibitem{abay}
Annie Abay, Yi~Zhou, Nathalie Baracaldo, Shashank Rajamoni, Ebube Chuba, and
  Heiko Ludwig.
\newblock Mitigating bias in federated learning, 2020.

\bibitem{raji2020closing}
Inioluwa~Deborah Raji, Andrew Smart, Rebecca~N White, Margaret Mitchell, Timnit
  Gebru, Ben Hutchinson, Jamila Smith-Loud, Daniel Theron, and Parker Barnes.
\newblock Closing the ai accountability gap: Defining an end-to-end framework
  for internal algorithmic auditing.
\newblock In {\em Proceedings of the 2020 conference on fairness,
  accountability, and transparency}, pages 33--44, 2020.

\bibitem{wieringa2020account}
Maranke Wieringa.
\newblock What to account for when accounting for algorithms: a systematic
  literature review on algorithmic accountability.
\newblock In {\em Proceedings of the 2020 conference on fairness,
  accountability, and transparency}, pages 1--18, 2020.

\bibitem{toreini2020relationship}
Ehsan Toreini, Mhairi Aitken, Kovila Coopamootoo, Karen Elliott,
  Carlos~Gonzalez Zelaya, and Aad Van~Moorsel.
\newblock The relationship between trust in ai and trustworthy machine learning
  technologies.
\newblock In {\em Proceedings of the 2020 conference on fairness,
  accountability, and transparency}, pages 272--283, 2020.

\bibitem{brundage_toward_2020}
Miles Brundage, Shahar Avin, Jasmine Wang, Haydn Belfield, Gretchen Krueger,
  Gillian Hadfield, Heidy Khlaaf, Jingying Yang, Helen Toner, and Ruth Fong.
\newblock Toward trustworthy {AI} development: mechanisms for supporting
  verifiable claims.
\newblock {\em {arXiv} preprint {arXiv}:2004.07213}, 2020.

\bibitem{bloomfield2020assurance}
Robin Bloomfield and John Rushby.
\newblock Assurance 2.0: A manifesto.
\newblock {\em arXiv preprint arXiv:2004.10474}, 2020.

\bibitem{kunnemann2019automated}
Robert K{\"u}nnemann, Ilkan Esiyok, and Michael Backes.
\newblock Automated verification of accountability in security protocols.
\newblock In {\em 2019 IEEE 32nd Computer Security Foundations Symposium
  (CSF)}, pages 397--39716. IEEE, 2019.

\bibitem{kunnemann2021accountability}
Robert Kunnemann, Deepak Garg, and Michael Backes.
\newblock Accountability in the decentralised-adversary setting.
\newblock In {\em 2021 IEEE 34th Computer Security Foundations Symposium
  (CSF)}, pages 1--16, 2021.

\bibitem{yumerefendi_trust_2004}
Aydan~R. Yumerefendi and Jeffrey~S. Chase.
\newblock Trust but verify: accountability for network services.
\newblock In {\em Proceedings of the 11th workshop on {ACM} {SIGOPS} European
  workshop}, pages 37--es, 2004.

\bibitem{kusters_accountability:_2010}
Ralf Küsters, Tomasz Truderung, and Andreas Vogt.
\newblock Accountability: definition and relationship to verifiability.
\newblock In {\em Proceedings of the 17th {ACM} conference on Computer and
  communications security}, pages 526--535. {ACM}, 2010.

\bibitem{wattenhofer2016science}
Roger Wattenhofer.
\newblock {\em The science of the blockchain}.
\newblock Inverted Forest Publishing, 2016.

\bibitem{kannengiesser2019does}
Niclas Kannengie{\ss}er, Sebastian Lins, Tobias Dehling, and Ali Sunyaev.
\newblock What does not fit can be made to fit! trade-offs in distributed
  ledger technology designs.
\newblock In {\em Proceedings of the 52nd Hawaii international conference on
  system sciences}, 2019.

\bibitem{greco2015datalog}
Sergio Greco and Cristian Molinaro.
\newblock Datalog and logic databases.
\newblock {\em Synthesis Lectures on Data Management}, 7(2):1--169, 2015.

\bibitem{androulaki2018hyperledger}
Elli Androulaki, Artem Barger, Vita Bortnikov, Christian Cachin, Konstantinos
  Christidis, Angelo De~Caro, David Enyeart, Christopher Ferris, Gennady
  Laventman, Yacov Manevich, et~al.
\newblock Hyperledger fabric: a distributed operating system for permissioned
  blockchains.
\newblock In {\em Proceedings of the thirteenth EuroSys conference}, pages
  1--15, 2018.

\bibitem{cachin2016architecture}
Christian Cachin et~al.
\newblock Architecture of the hyperledger blockchain fabric.
\newblock In {\em Workshop on distributed cryptocurrencies and consensus
  ledgers}, volume 310, pages 1--4. Chicago, IL, 2016.

\bibitem{ibmfl}
Heiko Ludwig, Nathalie Baracaldo, Gegi Thomas, Yi~Zhou, Ali Anwar, Shashank
  Rajamoni, Yuya Ong, Jayaram Radhakrishnan, Ashish Verma, Mathieu Sinn, et~al.
\newblock Ibm federated learning: an enterprise framework white paper v0. 1,
  2020.

\bibitem{whitepaperEUAI2020}
European Commission.
\newblock White paper on artificial intelligence: A european approach to
  excellence and trust.

\bibitem{walfish2015verifying}
Michael Walfish and Andrew~J Blumberg.
\newblock Verifying computations without reexecuting them.
\newblock {\em Communications of the ACM}, 58(2):74--84, 2015.

\end{thebibliography}

\end{document}